# Tiered Graph Autoencoders with PyTorch Geometric for Molecular Graphs

Daniel T. Chang (张遵)

*IBM (Retired)* dtchang43@gmail.com

**Abstract:**

Tiered latent representations and latent spaces for molecular graphs provide a simple but effective way to explicitly represent and utilize groups (e.g., functional groups), which consist of the atom (node) tier, the group tier and the molecule (graph) tier. They can be learned using the tiered graph autoencoder architecture. In this paper we discuss adapting tiered graph autoencoders for use with PyTorch Geometric, for both the deterministic tiered graph autoencoder model and the probabilistic tiered variational graph autoencoder model. We also discuss molecular structure information sources that can be accessed to extract training data for molecular graphs. To support transfer learning, a critical consideration is that the information must utilize standard unique molecule and constituent atom identifiers. As a result of using tiered graph autoencoders for deep learning, each molecular graph possesses tiered latent representations. At each tier, the latent representation consists of: node features, edge indices, edge features, membership matrix, and node embeddings. This enables the utilization and exploration of tiered molecular latent spaces, either individually (the node tier, the group tier, or the graph tier) or jointly, as well as navigation across the tiers.

## 1 Introduction

*Molecular graphs* [1] are widely used for representing molecular structure. They generally contain subgraphs (known as *groups*) that are identifiable and significant in composition, functionality, geometry, etc. *Tiered latent representations and latent spaces* for molecular graphs [2] provide a simple but effective way to explicitly represent and utilize groups, which consist of the *atom (node) tier*, the *group tier* and the *molecule (graph) tier*. They can be learned using the *tiered graph autoencoder architecture* [2].

*PyTorch Geometric (PyG)* [3] is a new library for deep learning on graphs, point clouds and manifolds. Following a simple *message passing API*, it bundles most of the recently proposed *convolution and pooling layers for graphs* into a single and unified framework.

In this paper we discuss adapting *tiered graph autoencoders* for use with PyG. The adaptation is for both the deterministic *tiered graph autoencoder (TGAE) model* and the probabilistic *tiered variational graph autoencoder (TVGAE) model*.

We also discuss molecular structure information sources that can be accessed to extract training data for molecular graphs. To support transfer learning, a critical consideration is that the information must utilize *standard unique molecule*

*and constituent atom identifiers*. This can be achieved by obtaining the source molecular structure information from *PubChem* through *ALATIS* as InChI and SDF files, which are accessed using *RDKit*.

## 2 Tiered Graph Autoencoders with PyG

The following diagram shows the *tiered graph autoencoder architecture* [2] for learning tiered latent representations and latent spaces:

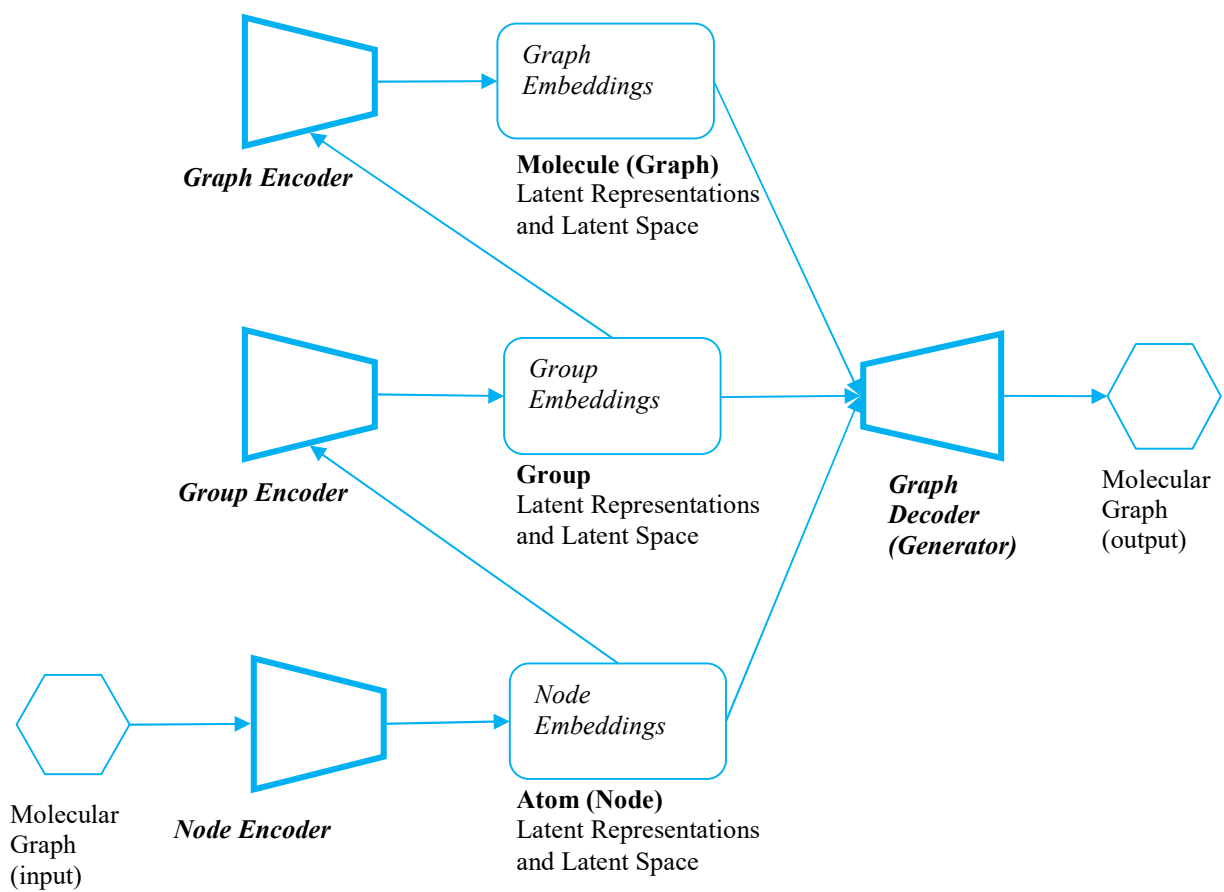

*Tiered graph autoencoders* use *graph neural networks (GNNs)*, *DiffGroupPool* and the *membership matrix* to learn and generate, jointly, node embeddings, group embeddings and graph embeddings [2].

In this paper, we adapt tiered graph autoencoders for use with PyG. The adaptation is for both the deterministic *TGAE model* and the probabilistic *TVGAE model*, which are discussed in the next two sections, respectively. Where needed, we map the notation and constructs used in tiered graph autoencoders to the notation, constructs and classes used in PyG. First, we unify the notation and constructs for graphs [2-3] as follows.

## 2.1 Graphs

Following PyG, a *graph* is represented as G = (**X**, (**I**, **E**)) where $\mathbf{X} \in R^{N \times d}$ is the *node feature matrix* (i.e., $\mathbf{F}^V$) and (**I**, **E**) is the *sparse adjacency tuple*. $\mathbf{I} \in N^{2 \times U}$ encodes *edge indices* in coordinate (COO) format and $\mathbf{E} \in R^{U \times s}$ is the *edge feature matrix* (i.e., $\mathbf{F}^E$).

A graph in PyG is described by an instance of the *Data* class, which has the following attributes: *x* (node feature matrix), *edge_index* (edge indices), *edge_attr* (edge feature matrix) as well as *pos* (node position matrix) and *y* (target). As an example, for the following undirected graph:

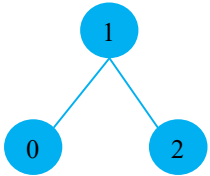

its edge_index contains the tensor [[0, 1, 1, 2], [1, 0, 2, 1]], which is more conveniently specified as a list of index tuples: [[0, 1], [1, 0], [1,2], [2,1]]. Note that although the graph has only two edges, we specify four index tuples to account for both directions of an edge. The *Data* object, when needed, may possess additional attributes, as we will see later.

The sparse adjacency tuple (**I**, **E**) is equivalent to the *dense adjacency matrix* $\mathbf{A} \in R^{N \times N \times s}$. (**I**, **E**) can be converted to **A** and vice versa. We will use (**I**, **E**) in general, but will occasionally use **A** where it is better suited, as is done in PyG.

## 3. TGAE Model
### 3.1 GNNs

Following PyG, we consider GNNs that employ the following *message passing* scheme [3-4] for node i at layer k:

$$\mathbf{x}_i^{(k)} = \gamma^{(k)}(\mathbf{x}_i^{(k-1)}, \Pi_j \varphi^{(k)}(\mathbf{x}_i^{(k-1)}, \mathbf{x}_j^{(k-1)}, \mathbf{e}_{ij}))$$

where j ∈ $\mathcal{N}(i)$ denotes a neighbor node of node i. $\mathbf{x}_i$ is the node feature vector and $\mathbf{e}_{ij}$ is the edge feature vector. γ and φ denote differentiable functions, e.g., MLPs. $\prod$ denotes a differentiable aggregation function, e.g., sum, mean or max.

A message-passing GNN is described in PyG by an instance of the *MessagePassing* base class, which automatically takes care of message propagation when the *propagate()* method is called. The user only has to define the function φ, i.e., the *message()* method, and γ, i.e., the *update()* method, as well as the aggregation scheme to use, i.e., the *aggr* parameter ("add", "mean" or "max"), and the direction of flow, i.e., the *flow* parameter ("source_to_target" or "target_to_source").

A full GNN will run K iterations to generate the output *node embedding* $z_i = x_i^{(K)}$, where K is usually in the range 2-6. For simplicity, we use

$$\mathbf{Z} = GNN(\mathbf{X}, (\mathbf{I}, \mathbf{E}))$$

to denote the embeddings generated using a GNN that implements K iterations of message passing. The embeddings could be node embeddings, group embeddings or graph embeddings.

For each tier t we thus have ($\mathbf{Z}^{(1)}$: node embeddings, $\mathbf{Z}^{(2)}$: group embeddings, and $\mathbf{Z}^{(3)}$: graph embeddings):

$$\mathbf{Z}^{(t)} = GNN(\mathbf{X}^{(t)}, (\mathbf{I}^{(t)}, \mathbf{E}^{(t)})).$$

### 3.2 Stacking GNNs and DiffGroupPool Modules

Given $\mathbf{Z} = GNN(\mathbf{X}, (\mathbf{I}, \mathbf{E}))$, we stack GNNs and DiffGroupPool (see the next subsection) modules as follows:

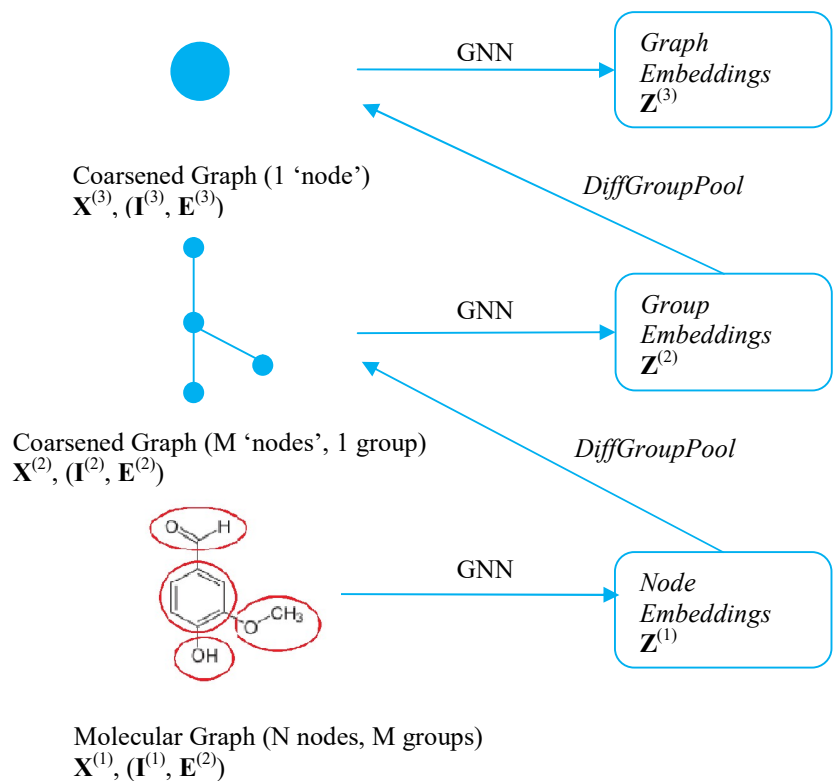

From the perspective of tiered graph autoencoders, we have the following *tiered encoders* with stacked GNNs and *DiffGroupPool* modules, from bottom to top:

- Node encoder: $\text{GNN}(X^{(1)}, (I^{(1)}, E^{(1)})) \Rightarrow Z^{(1)}$

- Group encoder: $\text{DGP}(Z^{(1)}, (I^{(1)}, E^{(1)}), M^{(1)}) \Rightarrow (X^{(2)}, (I^{(2)}, E^{(2)})), \text{GNN}(X^{(2)}, (I^{(2)}, E^{(2)})) \Rightarrow Z^{(2)}$

- Graph encoder: $\text{DGP}(Z^{(2)}, (I^{(2)}, E^{(2)}), M^{(2)}) \Rightarrow (X^{(3)}, (I^{(3)}, E^{(3)})), \text{GNN}(X^{(3)}, (I^{(3)}, E^{(3)})) \Rightarrow Z^{(3)}$

## 3.3 Differentiable Pooling via the Membership Matrix

The *DiffGroupPool* module for tier t+1 is denoted as:

$$(X^{(t+1)}, (I^{(t+1)}, E^{(t+1)})) = \text{DGP}(Z^{(t)}, (I^{(t)}, E^{(t)}), M^{(t)})$$

where $M^{(t)}$ is the *membership matrix* [2] at tier t.

It coarsens the input graph, generating a new coarsened sparse adjacency tuple $(I^{(t+1)}, E^{(t+1)})$ and a new matrix of coarsened embeddings $X^{(t+1)}$ in this coarsened graph. In particular, it applies the following equations:

$$\mathbf{X}^{(t+1)} = (\mathbf{M}^{(t)})^T \mathbf{Z}^{(t)} \in \mathbb{R}^{N\{t+1\} \times d}$$

$$\mathbf{A}^{(t+1)} = (\mathbf{M}^{(t)})^T \mathbf{A}^{(t)} \mathbf{M}^{(t)} \in \mathbb{R}^{N\{t+1\} \times N\{t+1\} \times s}$$

DiffGroupPool is identical to DiffPool [5] except that in the former the pooling is based on **M** whereas in the latter the pooling is based on the clustering assignment matrix **S**. DiffPool is supported in PyG by the *dense_diff_pool*() function (warning: it breaks when there are multiple edge features).

### 3.4 Graph Decoder (Generator)

For ease of adaptation for use with PyG, we decouple the graph decoder (generator) into three separate decoders (generators), one at each tier. We denote the *decoder (generator) at each tier* as, omitting the tier superscript for simplicity:

$$(\hat{\mathbf{X}}, (\hat{\mathbf{I}}, \hat{\mathbf{E}})) = \text{DEC}(\mathbf{Z}),$$

where $\hat{\mathbf{X}}$ is the reconstructed node features matrix and $(\hat{\mathbf{I}}, \hat{\mathbf{E}})$ is the reconstructed sparse adjacency tuple. At each tier, the model is trained by minimizing the *reconstruction loss function*:

$$\text{LOSS}((\hat{\mathbf{X}}, (\hat{\mathbf{I}}, \hat{\mathbf{E}})), (\mathbf{X}, (\mathbf{I}, \mathbf{E}))).$$

Learning takes place separately at each tier, proceeding from the lower tier to the higher tier, i.e., from the node tier to the group tier to the graph tier.

### 3.5 GAE

For simplicity, we adopt the *Graph Auto-encoder (GAE)* [6] for use at each tier, which generates embeddings **Z** using a *graph convolutional network (GCN)* [7] encoder:

$$\mathbf{Z} = \text{GCN}(\mathbf{X}, (\mathbf{I}, \mathbf{E})).$$

The GAE default decoder reconstructs the dense adjacency matrix $\hat{\mathbf{A}}$ as follows:

$$\hat{\mathbf{A}} = \sigma(\mathbf{Z}\mathbf{Z}^T),$$

where $\sigma(.)$ is the logistic sigmoid function.

In PyG, GAE and GCN are supported by the *GAE* and *GCNConv* classes, respectively.

# 4 TVGAE Model

In the *TVGAE model*, we extend the deterministic *TGAE model* into a probabilistic model by extending its components: GNNs, *DiffGroupPool* and graph decoder, to work together as *variational autoencoders* [8].

We denote the *encoder (inference) model* as [2]:

$$q_\varphi(\mathbf{Z} \mid \mathbf{X}, (\mathbf{I}, \mathbf{E})) = q_\varphi(\mathbf{Z}^{(1)} \mid \mathbf{X}, (\mathbf{I}, \mathbf{E})) \, q_\varphi(\mathbf{Z}^{(2)} \mid \mathbf{Z}^{(1)}) \, q_\varphi(\mathbf{Z}^{(3)} \mid \mathbf{Z}^{(2)})$$

## 4.1 GNNs

The parameters for the encoder (inference) model are specified using two GNNs, for convenience denoted as:

$$(\boldsymbol{\mu}_Z, \boldsymbol{\sigma}_Z) = \text{GNN}_\varphi(\mathbf{X}, (\mathbf{I}, \mathbf{E}))$$

where $\boldsymbol{\mu}_Z$ and $\boldsymbol{\sigma}_Z$ denote the *sufficient statistics* for the variational marginals. Note $\varphi$ is different for $\boldsymbol{\mu}_Z$ and $\boldsymbol{\sigma}_Z$.

For each tier t we thus have:

$$(\boldsymbol{\mu}_Z^{(t)}, \boldsymbol{\sigma}_Z^{(t)}) = \text{GNN}_\varphi(\mathbf{X}^{(t)}, (\mathbf{I}^{(t)}, \mathbf{E}^{(t)})).$$

## 4.2 Stacking GNNs and DiffGraphPool Modules

Same as the TGAE model, given the output of the GNNs, we stack GNNs and DiffGroupPool (see the next subsection) modules as follows:

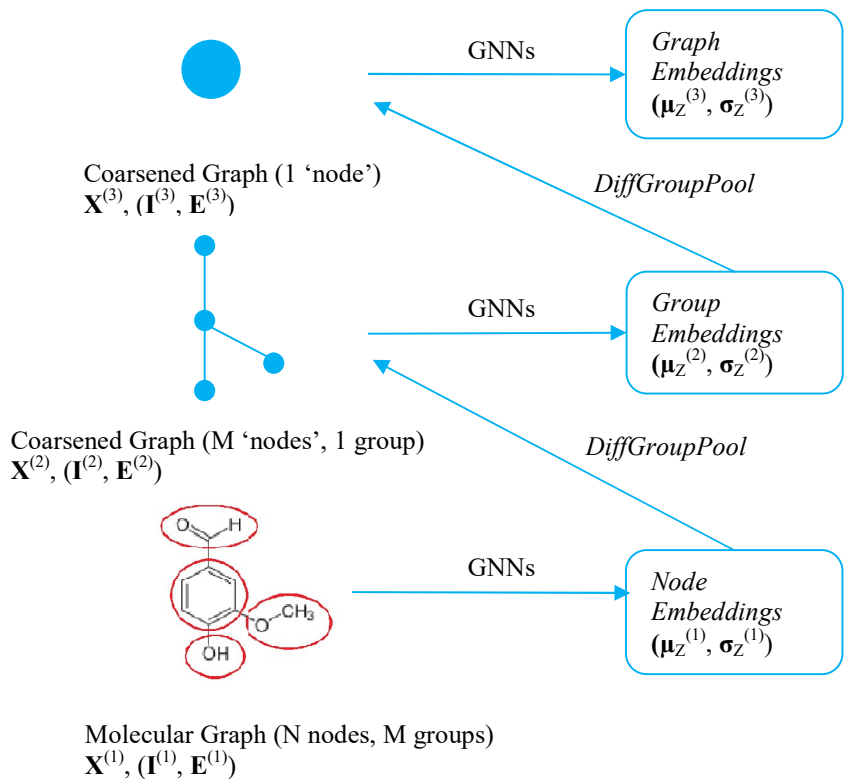

From the perspective of the TVGAE model, we have the following *tiered variational encoders* with stacked GNNs and *DiffGroupPool* modules, from bottom to top:

- Node encoder: $\text{GNN}_\varphi(\mathbf{X}^{(1)}, (\mathbf{I}^{(1)}, \mathbf{E}^{(1)})) \Rightarrow (\boldsymbol{\mu}_Z^{(1)}, \boldsymbol{\sigma}_Z^{(1)})$

- Group encoder: $\text{DGP}(\boldsymbol{\mu}_Z^{(1)}, (\mathbf{I}^{(1)}, \mathbf{E}^{(1)}), \mathbf{M}^{(1)}) \Rightarrow (\mathbf{X}^{(2)}, (\mathbf{I}^{(2)}, \mathbf{E}^{(2)})), \text{GNN}_\varphi(\mathbf{X}^{(2)}, (\mathbf{I}^{(2)}, \mathbf{E}^{(2)})) \Rightarrow (\boldsymbol{\mu}_Z^{(2)}, \boldsymbol{\sigma}_Z^{(2)})$

- Graph encoder: $\text{DGP}(\boldsymbol{\mu}_Z^{(2)}, (\mathbf{I}^{(2)}, \mathbf{E}^{(2)}), \mathbf{M}^{(2)}) \Rightarrow (\mathbf{X}^{(3)}, (\mathbf{I}^{(3)}, \mathbf{E}^{(3)})), \text{GNN}_\varphi(\mathbf{X}^{(3)}, (\mathbf{I}^{(3)}, \mathbf{E}^{(3)})) \Rightarrow (\boldsymbol{\mu}_Z^{(3)}, \boldsymbol{\sigma}_Z^{(3)})$

### 4.3 Differentiable Pooling via the Membership Matrix

Same as the TGAE model, the *DiffGroupPool* module for tier t+1 is denoted as (in this case we use $\boldsymbol{\mu}_Z$ in place of $\mathbf{Z}$):

$$(\mathbf{X}^{(t+1)}, (\mathbf{I}^{(t+1)}, \mathbf{E}^{(t+1)})) = \text{DGP}(\boldsymbol{\mu}_Z^{(t)}, (\mathbf{I}^{(t)}, \mathbf{E}^{(t)}), \mathbf{M}^{(t)})$$

It coarsens the input graph, generating a new coarsened sparse adjacency tuple $(\mathbf{I}^{(t+1)}, \mathbf{E}^{(t+1)})$ and a new matrix of coarsened embeddings $\mathbf{X}^{(t+1)}$ in this coarsened graph. In particular, it applies the following equations:

$$X^{(t+1)} = (M^{(t)})^T \mu_Z^{(t)} \in R^{N\{t+1\} \times d}$$

$$A^{(t+1)} = (M^{(t)})^T A^{(t)} M^{(t)} \in R^{N\{t+1\} \times N\{t+1\} \times s}$$

### 4.4 Graph Decoder (Generator)

For ease of adaptation for use with PyG, we decouple the graph decoder (generator) into three separate decoders (generators), one at each tier. We denote the *decoder (generator) model at each tier* as [8], omitting the tier superscript for simplicity:

$$p_\theta(X, (I, E) | Z)$$

Together with the encoder (inference) model at the corresponding tier, $q_\varphi(Z | X, (I, E))$, the model is trained using *variational learning* [8], which maximizes the *variational lower bound*:

$$L(X; \theta, \varphi) = \sum_Z q_\varphi(Z | X, (I, E)) \log(p_\theta(X, (I, E) | Z)) - D_{KL}(q_\varphi(Z | X, (I, E)) \| p_\theta(Z)).$$

Learning takes place separately at each tier, proceeding from the lower tier to the higher tier, i.e., from the node tier to the group tier to the graph tier.

### 4.5 VGAE

For simplicity we adopt the *Variational Auto-encoder (VGAE)* [6] for use at each tier. It uses an inference model:

$$q(Z | X, (I, E)) = \prod_{i=1}^{N} q(z_i | X, (I, E)), \text{ with } q(z_i | X, (I, E)) = \mathcal{N}(z_i | \mu_i, \text{diag}(\sigma_i^2)),$$

where $z_i$ is the latent variable for node i and $\mathcal{N}(.)$ is the Gaussian distribution. The inference model is parameterized by two GCNs:

$$\mu = GCN_\mu(X, (I, E)) \text{ and } \log\sigma = GCN_\sigma(X, (I, E)),$$

where $\mu$ is the matrix of $\mu_i$ and $\sigma$ is the matrix of $\sigma_i$.

The generative model is given by an inner product between latent variables:

$$p(\mathbf{A} \mid \mathbf{Z}) = \prod_{i=1}^{N} \prod_{j=1}^{N} p(\mathbf{A}(i,j) \mid \mathbf{z}_i, \mathbf{z}_j), \text{ with } p(\mathbf{A}(i,j) = 1 \mid \mathbf{z}_i, \mathbf{z}_j) = \sigma(\mathbf{z}_i^T \mathbf{z}_j),$$

where σ(.) is the logistic sigmoid function. It further takes a Gaussian prior:

$$p(\mathbf{Z}) = \prod_i p(\mathbf{z}_i) = \prod_i \mathcal{N}(\mathbf{z}_i \mid \mathbf{0}, \mathbf{I}).$$

The VGAE is trained by maximizing the variational lower bound.

In PyG, VGAE is supported by the *VGAE* class.

## 5 Molecular Structure Information Sources

The source molecular structure information is obtained from PubChem through ALATIS as InChI and SDF files, which are accessed using RDKit. To support transfer learning, a critical consideration is that the information must utilize *standard unique molecule and constituent atom identifiers*.

### 5.1 PubChem

*PubChem* (https://pubchem.ncbi.nlm.nih.gov/) is a chemical information repository. It is the world's largest collection of freely accessible chemical information (e.g., 96.5 million unique molecular structures). One can search chemicals by name, identifier, molecular formula, and molecular structure; and find physical and chemical properties, biological activities, safety and toxicity information, patents, literature citations and more.

PubChem consists of three inter-linked databases: Substance, Compound, and BioAssay [9-10]. Contributors submit chemical substance descriptions to PubChem, and the unique *molecular structures* are extracted and stored into the *Compound database* through an automated process called *structure standardization* [11]. This process consists of two major phases: *structure verification* and *structure normalization*, which can further be divided into four verification steps (*element, hydrogens, functional groups, and valence*) and five normalization steps (*annotations, valence bond, aromaticity, stereochemistry, and explicit hydrogens*).

As an example, the search result of *vanillin* in PubChem is summarized as:

- Compound CID: 1183
- MF: C8H8O3

- MW: 152.149g/mol
- InChIKey: MWOOGOJBHIARFG-UHFFFAOYSA-N
- IUPAC Name: 4-hydroxy-3-methoxybenzaldehyde

PubChem uses the IUPAC *InChI (International Chemical Identifier)* for molecular representation which is capable of assigning standard unique *molecule identifiers*. However, the current implementation of InChI fails to provide a complete standard for constituent atom nomenclature.

## 5.2 ALATIS

*ALATIS* (http://alatis.nmrfam.wisc.edu/) [12] is an adaptation of InChI, which operates fully within the InChI convention to provide standard and unique *molecule and constituent atom identifiers*. ALATIS includes an InChI extension for unique atom labeling of symmetric molecules.

The key to producing standard unique molecule and constituent atom identifiers is a *3D SDF file*, because it offers a complete representation of the molecule. Given the SDF file, ALATIS follows a strict protocol to derive the unique identifiers. This protocol consists of three main modules that (a) label heavy atoms, (b) label hydrogen atoms with special considerations for chiral, prochiral, and primary amide centers of symmetric and asymmetric molecules, and (c) label all atoms in the molecule.

ALATIS outputs a standard *InChI string* for the molecule, a *SDF file* that contains the unique identifiers of the constituent atoms, and a map between the atoms identifiers of the input and the generated unique atom identifiers.

As an example, the search result of *vanillin* in PubChem through ALATIS is summarized as:

- ALATIS formula: C8H8O3
- PubChem formula: C8H8O3
- PubChem weight: 152.149
- PubChem mass: 152.047
- Standard InChI: InChI=1S/C8H8O3/c1-11-8-4-6(5-9)2-3-7(8)10/h2-5,10H,1H3

## 5.3 SDF Format

Chemical table files formats (*CTfile formats*) [13] are used for representing and communicating chemical information, which is expressed in the form of connection tables. A *connection table (Ctab)* contains information describing the structural relationships and properties of a collection of atoms. Such collections might, for example, describe molecules, molecular fragments, substructures, substituent groups, polymers, alloys, formulations, mixtures, and even unconnected atoms. Multiple Ctabs can be included in an *SDF (Structure Data File)* file.

A CTab consists of an atom block, a bond block and other information. An *atom block* specifies all *node (atom) information* for the connection table. It specifies the atomic symbol and any mass difference, charge, stereochemistry, and associated hydrogens for each atom. A *bond block* specifies all *edge (bond) information* for the connection table. It specifies the two atoms connected by the bond, the bond type, and any bond stereochemistry and topology (chain or ring properties) for each bond.

As an example, the following information can be extracted for *vanillin* from the atom block and bond block of an SDF file:

- Nodes (atoms) - by identifier / index: [C, C, C, C, C, C, C, C, O, O, O, H, H, H, H, H, H, H, H]
- Edges (bonds): [[1, 12], [1, 13], [1, 14], [2, 3], [2, 15], [3, 16], [4, 17], [5, 18], [6, 2], [6, 4], [6, 5], [7, 3], [8, 4], [8, 7], [9, 5], [10, 7], [10, 19], [11, 1], [11, 8]]

Note that the node (atom) identifier / index starts at 1, not 0 as in PyG.

## 5.4 RDKit

*RDKit* (https://www.rdkit.org/) is an open source toolkit for cheminformatics [14]. It provides Python APIs for reading, writing and working with molecules. In particular, it provides the *ForwardSDMolSupplier* class for reading sets of molecules from an SDF file as *Mol* objects.

The *identify_functional_groups*() function (https://github.com/rdkit/rdkit/tree/master/Contrib/IFG) identifies the functional groups [15] in a Mol object, which is an important component for determining the membership matrix. The *mol2vec()* function (https://iwatobipen.wordpress.com/2019/04/05/make-graph-convolution-model-with-geometric-deep-

learning-extension-library-for-pytorch-rdkit-chemoinformatics-pytorch/) converts a RDKit Mol object into a PyG Data object for use in PyG.

# 6 Summary and Conclusion

Tiered latent representations and latent spaces for molecular graphs provide a simple but effective way to explicitly represent and utilize groups (e.g., functional groups), which consist of the atom (node) tier, the group tier and the molecule (graph) tier. They can be learned using the tiered graph autoencoder architecture.

In this paper we discussed adapting tiered graph autoencoders for use with PyG, for both the deterministic tiered graph autoencoder model and the probabilistic tiered variational graph autoencoder model. PyG proved to be an easy-to-use but powerful library for implementing tiered graph autoencoders. It supports both sparse adjacency tuple and dense adjacency matrix, with multiple edge features. It provides the key building blocks required for tiered graph autoencoders, including the Data, GCNConv, GAE and VGAE classes and the dense_diff_pool() function.

We also discussed molecular structure information sources that can be accessed to extract training data for molecular graphs. The source molecular structure information is obtained from PubChem through ALATIS as InChI and SDF files, which are accessed using RDKit. This ensures that the information utilizes standard unique molecule and constituent atom identifiers, and thus supports transfer learning.

As a result of using tiered graph autoencoders with PyG for deep learning, each molecular graph (a Data object) possesses tiered latent representations. At each tier, the latent representation consists of: node features, edge indices, edge features, membership matrix, and node embeddings. This enables the utilization and exploration of tiered molecular latent spaces, either individually (the node tier, the group tier, or the graph tier) or jointly, as well as navigation across the tiers.

For future work we plan to experiment with the tiered molecular latent spaces for molecular properties / activities prediction, as well as for molecular design. Other future work include: molecular graph generation using tiered embeddings (graph, group and node) with membership matrices, and extension of tiered graph autoencoders for 3D molecular graphs with configuration and conformation information.